\documentclass[pdflatex, sn-standardnature]{sn-jnl}

\jyear{2022}%
\theoremstyle{thmstyleone}%
\theoremstyle{thmstyletwo}%
\theoremstyle{thmstylethree}%
\usepackage{graphicx}
\usepackage{subcaption}
\usepackage{mwe}

\usepackage{glossaries}
\newacronym{MR}{MR}{magnetic resonance}
\newacronym{TTA}{TTA}{test-time augmentation}

\usepackage[most]{tcolorbox}
\definecolor{darkspringgreen}{rgb}{0.09, 0.45, 0.27}
\usepackage[frozencache,cachedir=mintfol]{minted} 

\begin{document}
\title[MONAI: An open-source framework for deep learning in healthcare]{MONAI: An open-source framework for deep learning in healthcare}

\author*[1]{\fnm{M. Jorge} \sur{Cardoso}} 
\equalcont{Equal contribution}
\author[2]{\fnm{Wenqi} \sur{Li}} 
\equalcont{Equal contribution}
\author[1]{\fnm{Richard} \sur{Brown}} 
\author[2]{\fnm{Nic} \sur{Ma}} 
\author[1]{\fnm{Eric} \sur{Kerfoot}} 
\author[2]{\fnm{Yiheng} \sur{Wang}} 
\author[1]{\fnm{Benjamin} \sur{Murrey}} 
\author[2]{\fnm{Andriy} \sur{Myronenko}} 
\author[2]{\fnm{Can} \sur{Zhao}} 
\author[2]{\fnm{Dong} \sur{Yang}} 
\author[2]{\fnm{Vishwesh} \sur{Nath}} 
\author[2]{\fnm{Yufan} \sur{He}} 
\author[2]{\fnm{Ziyue} \sur{Xu}} 
\author[2]{\fnm{Ali} \sur{Hatamizadeh}} 
\author[2]{\fnm{Andriy} \sur{Myronenko}} 
\author[2]{\fnm{Wentao} \sur{Zhu}} 
\author[2]{\fnm{Yun} \sur{Liu}} 
\author[2]{\fnm{Mingxin} \sur{Zheng}} 
\author[2]{\fnm{Yucheng} \sur{Tang}} 
\author[2]{\fnm{Isaac} \sur{Yang}} 
\author[2]{\fnm{Michael} \sur{Zephyr}} 
\author[2]{\fnm{Behrooz} \sur{Hashemian}} 
\author[2]{\fnm{Sachidanand} \sur{Alle}} 
\author[2,3]{\fnm{Mohammad} \sur{Zalbagi Darestani}} 
\author[1]{\fnm{Charlie} \sur{Budd}}  
\author[1]{\fnm{Marc} \sur{Modat}} 
\author[1]{\fnm{Tom} \sur{Vercauteren}} 
\author[1]{\fnm{Guotai} \sur{Wang}}  
\author[4]{\fnm{Yiwen} \sur{Li}}  
\author[4,5]{\fnm{Yipeng} \sur{Hu}}  
\author[5,6]{\fnm{Yunguan} \sur{Fu}}  
\author[7]{\fnm{Benjamin} \sur{Gorman}}  
\author[7]{\fnm{Hans} \sur{Johnson}} 
\author[2]{\fnm{Brad} \sur{Genereaux}}
\author[8]{\fnm{Barbaros S.} \sur{Erdal}} 
\author[8]{\fnm{Vikash} \sur{Gupta}} 
\author[1.2]{\fnm{Andres} \sur{Diaz-Pinto}}
\author[9]{\fnm{Andre} \sur{Dourson}} 
\author[10]{\fnm{Lena} \sur{Maier-Hein}} 
\author[11]{\fnm{Paul F.} \sur{Jaeger}} 
\author[12]{\fnm{Michael} \sur{Baumgartner}} 
\author[18]{\fnm{Jayashree} \sur{Kalpathy-Cramer}} 
\author[2]{\fnm{Mona} \sur{Flores}} 
\author[19]{\fnm{Justin} \sur{Kirby}} 
\author[20]{\fnm{Lee A.D.} \sur{Cooper}} 
\author[2]{\fnm{Holger R.} \sur{Roth}} 
\author[2]{\fnm{Daguang} \sur{Xu}} 
\author[2]{\fnm{David} \sur{Bericat}} 
\author[12,13]{\fnm{Ralf} \sur{Floca}} 
\author[14]{\fnm{S. Kevin} \sur{Zhou}} 
\author[15]{\fnm{Haris} \sur{Shuaib}} 
\author[16]{\fnm{Keyvan} \sur{Farahani}} 
\author[12,13]{\fnm{Klaus H.} \sur{Maier-Hein}} 
\author[17]{\fnm{Stephen} \sur{Aylward}} 
\author[2]{\fnm{Prerna} \sur{Dogra}} 
\author[1]{\fnm{Sebastien} \sur{Ourselin}} 
\equalcont{Equal contribution}
\author[2]{\fnm{Andrew} \sur{Feng}} 
\equalcont{Equal contribution}

\affil[1]{\orgdiv{School of Biomedical Engineering \& Imaging Sciences}, \orgname{King's College London}, \orgaddress{\city{London}, \country{U.K}}}
\affil[2]{\orgname{NVIDIA Corporation}, \orgaddress{\city{Santa Clara} and \city{Bethesda}, \country{USA}}}
\affil[3]{\orgdiv{School of Electrical and Computer Engineering}, \orgname{Rice University}, \orgaddress{\city{Houston}, \country{USA}}}
\affil[4]{\orgdiv{Department of Engineering Science}, \orgname{University of Oxford}, \orgaddress{\city{Oxford}, \country{UK}}}
\affil[5]{\orgdiv{Department of Medical Physics and Biomedical Engineering}, \orgname{University College London}, \orgaddress{\city{London}, \country{UK}}}
\affil[6]{\orgdiv{Department of BioAI}, \orgname{InstaDeep Ltd}, \orgaddress{\city{London}, \country{UK}}}
\affil[7]{\orgdiv{Department of Electrical \& Computer Engineering}, \orgname{University of Iowa}, \orgaddress{\city{Iowa City}, \country{USA}}}
\affil[8]{\orgname{Mayo Clinic}, \orgaddress{\city{Jacksonville}, \country{USA}}}
\affil[9]{\orgname{Mars Incorporated}, \orgaddress{\city{}, \country{USA}}}
\affil[10]{\orgdiv{Div. Intelligent Medical Systems}, \orgname{German Cancer Research Center}, \orgaddress{\city{Heidelberg}, \country{Germany}}}
\affil[11]{\orgdiv{Interactive Machine Learning Group}, \orgname{German Cancer Research Center}, \orgaddress{\city{Heidelberg}, \country{Germany}}}
\affil[12]{\orgdiv{Division of Medical Image Computing}, \orgname{German Cancer Research Center}, \orgaddress{\city{Heidelberg}, \country{Germany}}}
\affil[13]{\orgdiv{Pattern Analysis and Learning Group}, \orgname{ Department of Radiation Oncology}, \orgaddress{\city{Heidelberg University Hospital}, \country{Heidelberg}}}
\affil[14]{\orgdiv{School of Biomedical Engineering \& Suzhou Institute for Advanced Research}, \orgname{University of Science and Technology of China}, \orgaddress{\city{Suzhou}, \country{China}}}
\affil[15]{\orgdiv{Department of Medical Physics}, \orgname{Guy's \& St Thomas' NHS Foundation Trust}, \orgaddress{\city{London}, \country{UK}}}
\affil[16]{\orgname{National Cancer Institute}, \orgaddress{\city{Bethesda}, \country{USA}}}
\affil[17]{\orgname{Kitware}, \orgaddress{\city{ Inc.}, \country{Clifton Park}, \country{USA}}}
\affil[18]{\orgdiv{Department of Ophthalmology}, \orgname{ University of Colorado}, \orgaddress{\city{Aurora}, \country{USA}}}
\affil[19]{\orgname{Frederick National Laboratory for Cancer Research}, \orgaddress{\city{Frederick}, \country{USA}}}
\affil[20]{\orgdiv{Department of Pathology}, \orgname{Northwestern University}, \orgaddress{\city{Chicago}, \country{USA}}}

\abstract{Artificial Intelligence (AI) is having a tremendous impact across most areas of science. Applications of AI in healthcare have the potential to improve our ability to detect, diagnose, prognose, and intervene on human disease. For AI models to be used clinically, they need to be made safe, reproducible and robust, and the underlying software framework must be aware of the particularities (e.g. geometry, physiology, physics) of medical data being processed. This work introduces MONAI, a freely available, community-supported, and consortium-led PyTorch-based framework for deep learning in healthcare. MONAI extends PyTorch to support medical data, with a particular focus on imaging, and provide purpose-specific AI model architectures, transformations and utilities that streamline the development and deployment of medical AI models. MONAI follows best practices for software-development, providing an easy-to-use, robust, well-documented, and well-tested software framework.  MONAI preserves the simple, additive, and compositional approach of its underlying PyTorch libraries. MONAI is being used by and receiving contributions from research, clinical and industrial teams from around the world, who are pursuing applications spanning nearly every aspect of healthcare.}

\maketitle

\section{Introduction}\label{sec1}
Evidence-based medicine is the \textit{de facto} standard approach guiding clinical management and decision-making, drawing recommendations directly from robust and bias-controlled evidence, and utilizing statistical measures to quantify the clinical benefit of these recommendations. As patterns become more complex and data more numerous, advanced data analytics and pattern recognition methods are needed; artificial intelligence, and more specifically, machine learning, delivers on this need. 

Artificial intelligence (AI), when applied to healthcare data, carries the promise of improving and automating the detection and diagnosis of diseases, enabling the prognosis of disease outcomes, and supporting the delivery of clinical interventions. AI can also be used to speed image reconstruction, to automate clinical data curation and preparation \cite{Kraljevic:2021aa}, to improve auditing and clinical safety \cite{Brzezicki:2020aa}, and to optimise patient scheduling and other hospital operations \cite{Nelson:2019aa}. In order for this vision to be achieved, several problems need to be mitigated \cite{Panch:2019aa}; among them, the evidence supporting the safety and efficacy of these AI systems needs to outweigh the potential risks while representing a good value proposition. While several AI risks are often problem-specific and require substantial evidence to validate their accuracy, the safety and quality of the underlying software frameworks and tools used to develop these AI models are equally important and pervasive.  

There are a plethora of competing high-quality, general-purpose machine learning libraries and software frameworks that are used by researchers and developers to create, train, and deploy AI models. Examples of high-quality frameworks for AI development include actively supported frameworks such as Tensorflow \cite{tensorflowpaper}, Keras, PyTorch \cite{pytorchpaper}, JAX and Apache MXNet \cite{mxnetpaper}, as well as many deprecated frameworks such as Theano, Torch, Caffe, and CNTK. The two most popular frameworks (Tensorflow and PyTorch) have significant feature parity, demonstrating the maturity of the field. However, as general-purpose frameworks, they often do not support field-specific or data-specific functionality. Healthcare data, such as medical images and digital pathology reports, have a physiological basis that must be understood and maintained as the data is processed, e.g., tabular blood data transformations must be aware of the expected numerical ranges, or clinical x-ray computed tomography images are recorded in DICOM object files with rich metadata descriptions of the acquisition protocol and with voxel values recorded in Hounsfield units at specific within-slice and inter-slice distances. Correspondingly, the associated AI models for processing such data (e.g. 3D multimodal medical imaging data)  require purpose-specific architectures, augmentation, and training mechanisms. If healthcare AI models are to be built using a general-purpose framework, significant functionality would need to be developed and tested, thus increasing the length of and the risks associated with the full R\&D life-cycle. 

With the increasing popularity of AI in healthcare, several open-source healthcare-specific frameworks have also been developed by academic groups (NiftyNet \cite{Gibson:2018aa}, DLTK \cite{DLTKpaper}, DeepNeuro \cite{deepneuropaper}) and industry teams (NVIDIA Clara, Microsoft Project InnerEye). However, this disjointed development of several platforms resulted in a fragmented software field, diluting development efforts, reducing code quality, and slowing the pace of research. 

Project MONAI (https://monai.io) is a joint initiative aiming at defining, standardizing, developing, and exchanging best practices for AI in healthcare; it is unifying the fragmented healthcare AI software field. Initially led by NVIDIA and King's College London, Project MONAI has since evolved into a growing consortium currently with 16 different universities and industrial partners. Project MONAI consists of several components, including MONAI Label for AI-assisted image annotation, MONAI Deploy for integrating AI models into clinical workflows, and MONAI Core for deep learning model research and development.  This paper is focused on MONAI Core, but much of the design and community-oriented philosophies of MONAI Core are shared with the other Project MONAI components.

MONAI Core is a freely available, community-supported, consortium-led, component of Project MONAI for deep learning in healthcare. It provides domain-optimized foundational capabilities for developing healthcare AI model training workflows, with a focus on imaging, video, and other forms of structured data (e.g. tabular data, HL7 FIHR, EEG signals). MONAI Core has strong community backing, with several thousand active users and more than 120 contributors. It was created with several principles in mind: open-source design, standardized, user-friendly, reproducible, high quality, and easy to integrate. This paper describes the MONAI Core software framework and its open-source strategy, design philosophy, implementation choices, abstractions, and capabilities.  This paper also provides examples of how MONAI Core, and Project MONAI in general, is being applied to solve a variety of healthcare challenges. 

\subsection{Design considerations for deep learning in healthcare}
Compared with generic deep learning software packages, healthcare has its own characteristics and technical challenges. These aspects are carefully considered throughout the design of the project.
First, the data are often in the form of high-dimensional arrays, with the processing of the data often being coupled with the underlying physical interpretations of the data acquisition processes and the anatomical analysis. Thus, the design should consider the accompanying metadata and the relevant annotations.
Second, due to data variability and highly flexible requirements, it is necessary to build simple and robust low-level data processing components, with each component focusing on a key problem implemented to be efficient and tested to be robust; such an approach allows end-users to build complex and flexible compositional pipelines using these components.
Lastly, while exposing the low-level component APIs, the design decisions should also be aware of the widely-used, high-level workflows. The introduction of standardized pipeline elements reduces the effort needed to reproduce algorithmic research baselines, potentially accelerating research and development by allowing users to focus on their innovative contributions while reusing most of the existing pipelines.

\section{Methods}\label{sec11}
\subsection{Design choices and structure of the framework}
As alluded to in the previous section, MONAI Core (also referred to simply as MONAI in the remainder of this paper) has been through a very deliberate design process with a clear design philosophy as its foundation. Project MONAI has a set of principles to which its design adheres and which guide the addition of functionality across the core team and the broader base of contributors.

The three key design principles are: it looks and feels like PyTorch; it is opt-in and incremental over PyTorch, and it fully integrates with the PyTorch ecosystem. These principles are detailed in the following sections, with concrete examples throughout subsequent sections.

\subsubsection{MONAI looks and feels like PyTorch}
PyTorch is a highly-successful deep learning framework, particularly amongst the research community \cite{he2019mlframeworks}. This research focus was a key factor in why PyTorch was chosen as the MONAI underlying framework, with ease of use when building novel networks and a rich ecosystem built up around the core PyTorch functionality, further motivating this decision.

MONAI has adopted the principle that features should be designed and implemented following the core guidelines and principles of PyTorch (\url{docs.monai.io/en/stable/contrib.html}), or in the case of the quality-of-life frameworks, such as Ignite (\url{pytorch.org/ignite}), as if it was done by the team developing that library. This approach was chosen to ensure developers using MONAI have a minimal learning curve and cognitive burden beyond understanding PyTorch.

Exceptions to this rule are additive in nature. As an example of such an exception, a number of MONAI transforms which rely on a random number generation have a behaviour typical of transforms provided by libraries such as Torchvision (\url{pytorch.org/vision}), the de-facto imaging preprocessing library for PyTorch, but expose extra options that can provide additional functionality to enhance determinism.

\subsubsection{MONAI is opt-in and incremental over PyTorch}
A core tenet of MONAI is to be opt-in and incremental, and emphases low coupling between components. Rather than rebuilding a network architecture or model in MONAI, individual components can be selected to enhance vanilla PyTorch, Ignite-based projects, or other deep learning pipelines. MONAI has a smooth barrier-to-entry, allowing the user to gradually and seamlessly adopt individual transforms, layers, loss functions, and other elements of a deep learning model into a vanilla PyTorch model.

In particular, MONAI does not enforce overly stringent design philosophies or offer cumbersome APIs, nor does its implementation use architectural features that are unfamiliar to or introduce unexpected constraints on Pytorch users.
MONAI definitions are almost entirely standard Python, inheriting from PyTorch or Ignite types and exhibiting the same loose coupling expected of these libraries.
Only PyTorch, and by extension Numpy, are mandatory dependencies of MONAI; if libraries, such as Ignite, are absent from the environment, then dependent definitions will use alternative or dummy implementations with appropriate error management.

\subsubsection{MONAI collaborates with the PyTorch ecosystem}
There is a wealth of medically-focused libraries built around PyTorch, which are seldom compatible with one another. For example, while there is a rich ecosystem of medical transform and augmentation libraries, combining these libraries (e.g., through PyTorch Compose) is not directly supported. In collaboration with the developers of these libraries, MONAI now allows methods from those libraries to be mixed within a single compose command, promoting their usage and integrating their benefits and user communities.

MONAI also integrates and seamlessly extends several core PyTorch ecosystem tools. As an example, several PyTorch-based quality-of-life frameworks, such as Ignite, exist to provide higher-order engines and abstractions for the training and validation of neural networks. MONAI has adopted Ignite and extended it following an Ignite-native approach, all while ensuring two-way compatibility.

\subsection{Open-Source Strategy}
MONAI is a software framework for both research and commercial product development, with the ultimate objective of accelerating, simplifying, and reducing risks associated with AI model development and associated downstream clinical deployment. With this in mind, MONAI is licensed under the Apache-2.0 License, a permissive, altruistic, free-software license that allows the software to be used for any purpose, to be distributed and modified, without Project MONAI charging royalties or licensing fees. Copyleft licenses were explicitly avoided as the redistribution requirement would result in reduced commercial adoption and contributions, which have significantly benefited the quality, capabilities, and impact of MONAI. 

While MONAI is being developed with minimal required dependencies, namely PyTorch and NumPy, it provides wrappers and adaptors that allow popular healthcare AI tools to be used from within MONAI. Such an approach improves access to these tools and brings the community together in a manner that preserves software quality and integration. 

Several other software initiatives are being developed under the Project MONAI consortium, building on the foundation provided by MONAI Core: MONAI Label, an AI-assisted data labeling tool \cite{MONAILabel}; MONAI Deploy, a framework and software development toolkit allowing AI models to be deployed and integrated into a clinical environment; MONAI FL, providing standards for Federated Learning; MONAI Education, an educational package that can be used or adapted to teach AI in healthcare; with others on the way. 
The Project MONAI consortium is also working with key societies in the field, such as MICCAI and MIDL, with the aim of supporting educational initiatives and hackathons and bringing the research community together.

\subsection{System Overview and components}

\begin{figure}[b!]
	\begin{center}
	\includegraphics[width=\textwidth]{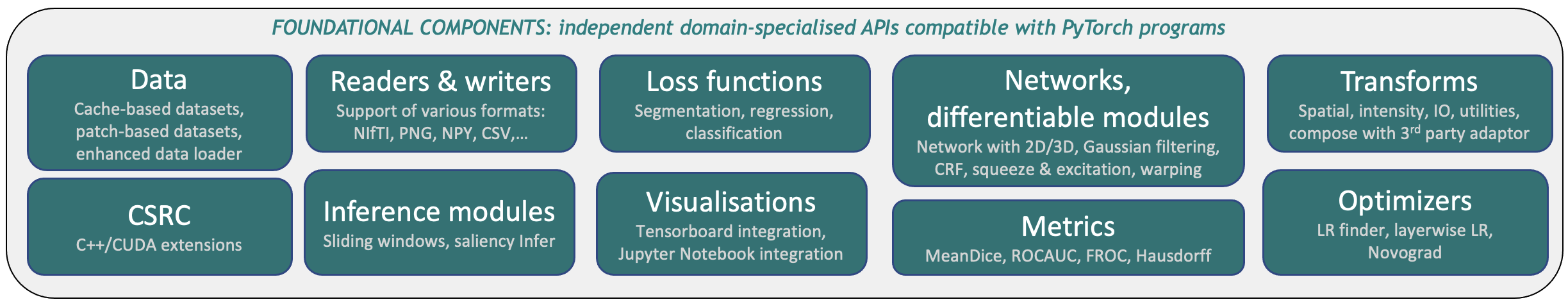}
    \includegraphics[width=\textwidth]{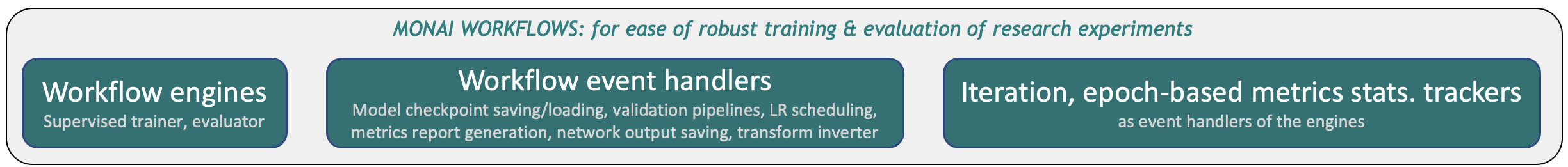}
	\end{center}
	\caption{Pictorial representation of the MONAI Core modules (top) and MONAI workflows components (bottom). }
	\label{modules}
\end{figure}

The MONAI core modules include the following (as illustrated in Fig \ref{modules}-Top): 
\texttt{monai.data}, targeting the datasets, readers/writers, and synthetic data;
\texttt{monai.losses}, targeting classes defining loss functions and following the pattern of \texttt{torch.nn.modules.loss};
\texttt{monai.networks}, containing network definitions, component definitions, and PyTorch-specific utilities;
\texttt{monai.transforms}, defining data transforms for preprocessing and postprocessing;
\texttt{monai.csrc} and \texttt{monai.\_extensions}, providing {C++/CUDA} extensions for MONAI core Python APIs;
\texttt{monai.visualize}, containing utilities for data visualization;
\texttt{monai.metrics}, defining metric tracking and analysis tools;
and \texttt{monai.optimizers}, containing classes defining optimizers and following the pattern of \texttt{torch.optim}.
To assemble and integrate different components, as depicted in Fig \ref{modules} (bottom), core modules also include \texttt{monai.engines} and \texttt{monai.handlers}, defining workflow engines for training and evaluation, and event handlers for implementing functionality at various stages in the training and inferencing processes.

\subsection{Transforms}
Data transformation and augmentation is of paramount importance in the domain of deep learning, primarily when applied to imaging data. They can help with data processing, from simple use cases such as loading an image from a file or normalizing the intensity of an image to a given range, to complex geometrical data transformations taking the image geometry into account. Equally, transforms can be used in the context of data augmentation to minimize model overfitting, for example, by randomly rotating or flipping images.

MONAI aims to support users by providing both domain-specific transforms (see next sections), as well as those commonly used in volumetric medical-imaging deep learning pipelines. Medical images require highly specialized methods for I/O, pre-processing, and augmentation. Images are often stored in complex formats with rich meta-information, with the data volumes being high-dimensional and requiring carefully designed manipulation procedures.

\begin{figure}[!b]
        \centering
        \begin{subfigure}[b]{0.475\textwidth}
            \centering
            \includegraphics[width=\textwidth]{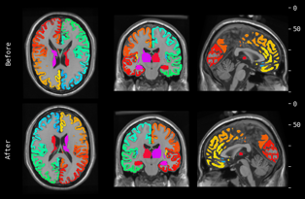}
            \caption[]%
            {{\small \texttt{Orientation} (RAS)}}    
            \label{fig:Orientation)}
        \end{subfigure}
        \hfill
        \begin{subfigure}[b]{0.475\textwidth}  
            \centering 
            \includegraphics[width=\textwidth]{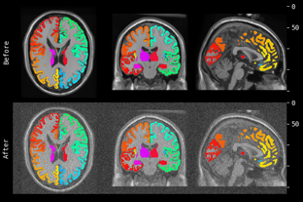}
            \caption[]%
            {{\small \texttt{RandGaussianNoise}}}    
            \label{fig:RandGaussianNoise}
        \end{subfigure}
        \begin{subfigure}[b]{0.475\textwidth}   
            \centering 
            \includegraphics[width=\textwidth]{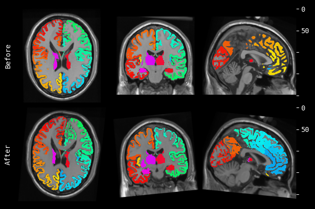}
            \caption[]%
            {{\small \texttt{RandRotate}}}    
            \label{fig:RandRotate}
        \end{subfigure}
        \hfill
        \begin{subfigure}[b]{0.5\textwidth}   
            \centering 
            \includegraphics[width=\textwidth]{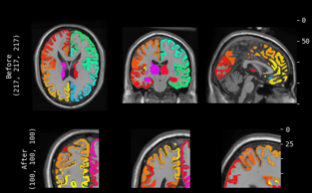}
            \caption[]%
            {{\small \texttt{RandSpatialCrop}}}    
            \label{fig:RandSpatialCrop}
        \end{subfigure}
        \caption[]
        {\small Each block (a-d) represents some data before (top) and after (bottom) a transformation.}
        \label{fig:transforms}
    \end{figure}

MONAI provides comprehensive medical image-specific transformations for IO, spatial, intensity, crop/pad, etc. They can be easily composed into optimized medical data processing pipelines emphasizing reproducible results. A non-exhaustive list of examples is provided as follows: \texttt{LoadImage}, which loads medical images with an automatically-selected package (\texttt{ITK}~\cite{itk2014}, \texttt{Nibabel}~\cite{nibabel}, \texttt{PIL}~\cite{PIL}, etc.); \texttt{Spacing}, which resamples input images into a specified voxel spacing; \texttt{Orientation}, which transforms images to a specified orientation axis (e.g., RAS); \texttt{RandGaussianNoise}, which perturbs image intensities by adding statistical noise; \texttt{NormalizeIntensity}, which normalizes data intensity based on mean and standard deviation; \texttt{Affine}, which transforms images based on affine parameters; and \texttt{Rand3DElastic}, which performs both a random elastic and affine deformations in 3D. Some of these examples are also depicted in Figure~\ref{fig:transforms}.

\subsubsection{Physics-specific transforms}

In a medical imaging setting, users often require transforms that have a grounding in the underlying physics of the acquisition. For example, data acquired from \gls*{MR} imaging is recorded in $k$-space, and therefore it is desirable to be able to augment \gls*{MR} images in the $k$-space domain. The \texttt{RandKSpaceSpikeNoise} transform, which randomly applies localized radio-frequency spikes in $k$-space, is just one example of many transforms in MONAI that are inspired by the physics of how medical images are acquired.

\subsubsection{Invertible transforms}\label{inverse_transforms}

Within deep learning workflows, it is often desirable to invert or revert previously-applied spatial transforms (resize, flip, rotate, zoom, crop, pad, etc.) in order to return transformed data to a previous geometrical state. This is useful if, for example, in \gls*{TTA}~\cite{testtimeaugmentation2019}, augmentation-consistency based domain adaptation, or when the user wishes to save an inferred segmented image in a way that preserves the original geometry and promotes  transformation equivariance. Doing so requires a full understanding of the tensor geometry and attached rasterization grid as to ensure the geometry is preserved after inversion.

\subsubsection{Array and dictionary transforms}\label{transforms_array_dict}

The design of MONAI transforms emphasizes code readability and usability. MONAI transforms can either be applied to a tensor or array of data (e.g., a single image), or to a data dictionary. The former focuses on simplicity, whilst the latter allows for more complex pipelines, such as applying the same random instantiation of a transform to sets of paired data (i.e. input image and associated ground truth segmentation). Storing data in a dictionary also allows for meta-information to be used, updated and tracked, for example, to ensure that the physical attributes of an image are preserved after a resize transform is applied.

\subsubsection{Utility transforms}

The ability to compose and chain transforms is naturally required when applying a series of data processing steps; this is achieved in MONAI using the \texttt{Compose} class. The provided \texttt{OneOf} transform allows the user to build on \texttt{Compose} by executing one of multiple transforms with some user-defined probability.

To convert the data shapes or types, utility transforms such as \texttt{ToTensor}, \texttt{ToNumpy}, \texttt{SqueezeDim} are also provided. These are not only convenient for users but also enable compatibility with other packages, as described in the following sections.

\subsubsection{Simplifying CPU, GPU, PyTorch \& NumPy interactions}

The majority of transforms in MONAI are compatible with both PyTorch and NumPy input, do not require explicit conversion between the two formats, and  support both CPU and GPU inputs. Users can also arbitrarily switch between CPU and GPU devices during their chain of transforms with the \texttt{ToDevice} transform, allowing for computation-heavy transforms (e.g. those requiring image resampling) to be performed on the GPU, with other transforms performed on the CPU to save GPU memory for model training.

\subsubsection{Compatibility with community-led libraries}

MONAI provides adapter tools to accommodate the use of third-party transforms from packages such as \texttt{ITK}~\cite{itk2014}, \texttt{torchIO}~\cite{perez-garcia_torchio_2021}, \texttt{Kornia}~\cite{eriba2019kornia}, \texttt{BatchGenerator}~\cite{isensee_fabian_2020_3632567}, \texttt{Rising}~\cite{rising}, and \texttt{cuCIM}~\cite{cuCIM}. 

For example, \texttt{cuCIM} has implementations of optimized versions of several common transforms that are often used in digital pathology pipelines. These transforms are executed on the GPU and act on \texttt{CuPy} arrays, with the \texttt{CuCIM} class provided to adapt \texttt{cuCIM} operations to MONAI transform types. 

\subsection{Engines, Loss Functions, and Metrics}

MONAI focuses on PyTorch-like implementations, and thus, introducing additive concepts on top of its architecture allows its components to be used in typical existing workflows. The common pattern for a PyTorch training loop, consisting of an explicit iteration over input data which is passed to the model and loss calculation before the optimizer is manually stepped, can incorporate MONAI networks, loss function, metrics, and other features. Following our design principles, MONAI training and inference workflows extend the \texttt{Engine} class of Ignite. These types encapsulate the training loop while supporting online metric calculation, visualization, and saving network state. Functionality is added on top of that provided by Ignite, such as the support for specialized transform handling, provide default training functions, and provide extra event handling functionality.

Loss functions in MONAI adhere to PyTorch API conventions and can be used in any existing PyTorch workflow.
The added functions introduce specialized losses not present in the base library, such as Dice and its variants (generalized, masked, focal, etc.) \cite{Sudre_2017}, focal loss, Tversky loss, contrastive loss, those used for image registration (bending energy, multi-scale loss, etc.), and several others. Where possible, inheritance is used to limit code duplication and ease the extension of the provided classes, and many aspects of the loss calculations, such as choice of reduction, are controlled via constructor arguments.
In addition to PyTorch compatibility, the design of the loss functions emphasizes flexibility of use and ease of extension.

Lastly, metrics are used to calculate values for assessing the performance of an AI model during training or validation. These are typically non-differentiable, unlike loss functions, allowing them to calculate measures of performance that would be unsuitable as losses. MONAI has been among the lead initiatives of the \textit{Metrics Reloaded} consortium, comprising academic and industrial image analysis thought leaders from all over the world~\cite{MaierHein2022}. The mission of Metrics Reloaded is to foster reliable algorithm validation through problem-aware choice of metrics with the long-term goal of (1) enabling the reliable tracking of scientific progress and (2) bridging the current chasm between AI research and translation into biomedical imaging practice. Contributions include a comprehensive collection of metric-related pitfalls~\cite{Reinke2021} and a corresponding paper on recommendations~\cite{MaierHein2022}. Reference implementations for all recommended metrics are currently being integrated into MONAI. Following the metric definitions provided by Ignite, MONAI metric classes can be used with an evaluator engine class to assess the network at key points in training while supporting distributed data-parallel operations.

\subsection{Network Architectures}

\begin{table}[]
    \centering
    \begin{tabular}{l l l}
         \multicolumn{2}{c}{\textbf{Reference}} & \textbf{General Purpose} \\
         \hline\hline
            AHNet        & SegResNet    & AutoEncoder          \\
            BasicUNet    & SegResNetVAE & Regressor            \\
            DenseNet     & SENet        & Classifier           \\
            DiNTS        & Transchex    & Discriminator        \\
            DynUNet      & UNETR        & Critic               \\
            EfficientNet & ViT          & FullyConnectedNet    \\
            HighResNet   & ViTAutoEnc   & VarFullyConnectedNet \\
            RegUNet      & VNet         & Generator            \\
            ResNet       & SwinUNETR    & UNet                 \\
                         &              & VarAutoEncoder       \\
    \end{tabular}
    \caption{MONAI Network Definitions}
    \label{tab:networks}
\end{table}

Table \ref{tab:networks} lists a subset of the network definitions provided by MONAI and divided into two primary categories: those defining a reference implementation of a published architecture, and those defined for general use following what is considered best design and practice. 
Reference implementations include networks such as \texttt{ResNet}, \texttt{BasicUNet}, \texttt{EfficientNet}, and several transformer-based networks, with implementations explicitly adhering to their published definitions and settings, rather than providing an abstract architecture. Where possible, these are also defined with some configurability in place, such as allowing 1/2/3D inputs and outputs based on constructor parameters, and a varying number of input and output channel counts.

General purpose networks, conversely, primarily adhere to an architectural structure emphasizing inheritance for reuse, configurability of number of spatial dimensions for input and output data, of internal structure (e.g., the number of levels or layers), and of specific network layers and blocks. The intent of this class of networks is to define sensible defaults, inheriting structure or components from other similar networks, while permitting customization and reuse in other applications without requiring major rewriting or adaptation.

\subsubsection{Working network example and extensions}
As an example, the \texttt{UNet} class can be instantiated without modification to operate on one, two, or three-dimensional data as input and output. 
The number of levels the network has can also be set to values other than that found in the reference implementation \cite{Ronneberger2015}, subject to the constraints of downsampling (as discussed later).
Each level is by default defined using blocks of convolution-normalization-dropout-activation layers, but these can be replaced with residual units or sequences of such layers.
Constructor arguments also set the type of activation or normalization used throughout the network.
As an explicit example, a network segmenting 3D volumetric image data into four labels can be created by setting the constructor parameters for \texttt{dimensions} to 3, \texttt{in\_channels} to 1 for single-channel inputs, and \texttt{out\_channels} to 4. 
The number of layers is set by the \texttt{channels} parameter whose values define the number of output channels for the layers of the encoding side of the network, \textit{e.g.,} five layers, comprising four encoding layers and a final bottleneck layer, would be given as \texttt{(4, 8, 16, 32, 64)}.

By default, downsampling is performed in the encoding part of the network using strided convolutions, as defined by the \texttt{strides} parameter for every layer save the bottleneck. 
A \texttt{strides} value of \texttt{(2, 2, 2, 2)} implies the network will downsample the activation map by half at each of the first four layers. Note that an input whose spatial dimensions are multiples of $2^4$ is necessary for the decoding path to concatenate skip connections and get an output with the same dimensions as the input. 

A much simpler network producing a single channel 2D output from the same sort of input can be created by changing the \texttt{dimensions} argument to 2 and \texttt{out\_channels} to 1. Using fewer layers can be accomplished with fewer values in \texttt{channels} and \texttt{strides}.
Activation layers are uniformly the same type throughout the network, by default PReLU \cite{https://doi.org/10.48550/arxiv.1502.01852}, but can be changed by setting the \texttt{act} argument to a name or a type for a different layer.
MONAI provides factory objects to query types from a central location by names, thus setting \texttt{act} to the string ``GELU''~\cite{gelu} will cause the network to be constructed with that activation in place of the default.

All general-purpose networks are compatible with Torchscript as scripted objects when using either the default or a subset of the layer parameters, maximizing portability and usability of pre-trained networks. This allows trained networks to be loaded in a process where there is no Python dependency, thus facilitating downstream deployment and integration. A scripted network saved to a Torchscript archive can be instantiated and used for training or inference without MONAI being installed in the host environment, thus decoupling the library and its version from the stored network so that requirements and incompatibilities with the host environment can be minimized.

These network factories are used to easily refer to built-in PyTorch or custom types, as well as choose the appropriate type for the desired spatial dimensionality. This latter property is useful for PyTorch layers, which define separate classes for each dimension as the factory can be queried by name and other arguments such as dimension. Without this feature, definitions parameterized on number of dimensions would have to include code to choose which layer to instantiate based on that value.

Typically MONAI definitions using these factories permit constructor argument values to be passed along with the name so that the layer can be constructed in a configurable way. For example, the \texttt{act} argument for \texttt{UNet} can be given \texttt{(Act.SELU, {"inplace": True})} to use \texttt{torch.nn.SELU} as an in-place operation \cite{https://doi.org/10.48550/arxiv.1706.02515}.

\subsection{Datasets and IO}

\subsubsection{Data representations for imaging}

Medical imaging data requires metadata to preserve geometric and other data-specific, information. As such, MONAI provides first-class support for imaging data and related networks and transforms. A typical approach to implementing such domain-specific functionality is to provide concrete types for entities such as medical images that are then hardwired through the framework. This, however, conflicts the MONAI design goal that it should be interoperable with PyTorch; it also limits the reusability of components in cases where the assumptions made about such specialized types do not hold.

To this end, MONAI has a class named \texttt{MetaTensor}. It inherits from the PyTorch \texttt{torch.Tensor} class, meaning it can be used wherever tensors are used in PyTorch. It further extends upon PyTorch functionality by also storing image metadata. An example of this is the orientation information stored in DICOM or NIfTI images, which when loaded with MONAI can then be stored alongside the images, and the MONAI transforms can use and update this meta data as necessary.

The \texttt{MetaTensor} also stores information, such as transformations that have already been applied to an image. This is particularly useful when data augmentations are only applied some of the time, parameterized by user-defined probabilities (e.g., randomly flipping an image 50\% of the time), or where the user would like to inspect the parameters of a randomly applied transform (e.g., knowing the rotation matrix that was applied during \texttt{RandRotate}). It also enables MONAI to perform actions such as inverse transforms (as mentioned in~\ref{inverse_transforms}) all whilst resting in the confines of PyTorch interoperability.


\subsubsection{Extended dataset functionality}

Medical datasets, particularly those for various imaging modalities, are amongst the most memory-intensive per-sample datasets in deep learning. Conversely, the relative lack of high-quality labeling means that, at least for supervised deep learning tasks, training must often be carried out on relatively small number of samples and data augmentation techniques are very important for effective learning. As a result, pre-processing of data can be a significant overhead unless careful attention is paid to the data input and pre-processing aspects of model training.

MONAI provides extensions to the standard PyTorch Dataset class which help solve the above problems. In particular, MONAI provides integrated caching and persistence solutions that can reduce the computational expense of data pre-processing without adding complexity to the user experience.

\texttt{CacheDataset} provides the ability to cache deterministic operations in a pre-processing pipeline that can be costly with certain large types of medical data. Operations such as deterministic pixel/voxel resampling and rescaling are carried out the first time when a particular sample is loaded, and the results of those operations are then cached \emph{in memory} for future training steps. Note that only preprocessing steps occurring before stochastic steps can be cached in this way. CacheDataset provides effective performance gains as long as the cached data does not exceed RAM, and the user can tune the proportion of the dataset that is cached in this manner.

\texttt{PersistentDataset} provides analogous functionality to \texttt{CacheDataset}, but with the output of the deterministic stages stored in an intermediary file system representation rather than to memory. The use of \texttt{PersistentDataset} is recommended for 3D datasets or where the overall dataset size is much larger than RAM. 

\subsubsection{Wrappers for reference datasets}
As with machine learning in other domains, reference datasets exist for medical deep learning applications. For example, MedNIST \footnote{Made available by Dr. Bradley J. Erickson M.D., PhD (Department of Radiology, Mayo Clinic) under the Creative Commons CC BY-SA 4.0 license.} serves as an MNIST handwritten digit dataset analogue \cite{lecun2010mnist}, providing a standard dataset that is suitable for initial training and benchmarking of various medically-focused deep learning applications. Another example, the Medical Segmentation Decathlon~\cite{decathlon}, represents a set of 10 challenge datasets, widely used to demonstrate network performance across a range of segmentation tasks. MONAI provides Dataset extension classes that simplify the downloading, storage and partitioning of the MedNIST and Medical Segmentation Decathlon datasets, and demonstrate a best practice for wrapping other such datasets. These datasets are extensions of MONAI CacheDataset, facilitating rapid training. In addition to providing an extended dataset class for the Medical Segmentation Decathlon, the MONAI Consortium also manages a \texttt{msd-for-monai} S3 bucket in US and EU regions, via AWS Open Data, with the aim of facilitating cloud-based model training and data distribution. A plan is currently in place to further extend beyond the current dataset pool and provide a common platform for large research data sharing and ingestion for the community. Recently MONAI took a first step towards this by introducing \texttt{TciaDataset} to automatically download and extract publicly-accessible datasets from The Cancer Imaging Archive with accompanying DICOM segmentations, and act as PyTorch datasets to generate training, validation, and test data.

\subsection{Training, Inference Engines and Event Handlers}
To quickly set up training and evaluation experiments, MONAI provides a set of workflows to significantly simplify the setup of a MONAI script and allow for fast prototyping. 

These features decouple the domain-specific components and the generic machine-learning processes. They also provide a set of unified APIs for higher-level applications (such as AutoML, and Federated Learning). The trainers and evaluators of the workflows are compatible with the Ignite engine and event-handler mechanism. There are rich event handlers in MONAI to independently attach to the trainer or evaluator, and users can register additional custom events to workflows.

Distributed data-parallel is an important feature of PyTorch to connect multiple GPU devices, on single or multiple nodes, to train or evaluate models. The distributed data-parallel APIs of MONAI are compatible with the native PyTorch distributed module, Ignite distributed module, Horovod, XLA, and SLURM.

For model inferences on large volumes, the sliding window approach is a popular choice to achieve high performance while having flexible memory requirements. A typical process comprises the selection of continuous windows on the original image, followed by iteratively running batched window inferences until all windows are analyzed. The inference outputs are then aggregated into a single segmentation map, and the results are saved to file or compute some evaluation metrics. This approach also supports `overlap` and `blending\_mode` configurations to better handle window overlap, resulting in improved performance.

The Ignite library provides event handlers to decouple training/inference pipelines from additional logic which gets triggered at set times or events during execution.
Users can extend handlers for specific requirements, use pre-existing handlers for logging, metric calculation, or other tasks, or attach their own defined handler routines. 
These will be triggered by specific events during execution such as at the completion of a batch or an epoch, or during data loading, allowing data to be analyzed, model parameters adjusted, or other operations at these key moments.
\subsection{Visualisations and utilities}

\subsubsection{Tensorboard}

When developing large workflows, it is crucial that users are able to visualize, track and understand processes and progress. To this end, MONAI provides many helper functions to the popular package, Tensorboard~\cite{abadi2016tensorflow}. Images in medical imaging are often 3D, so MONAI wrapper functions enable the presentation of these images in the form of GIFs, a set of 2D slices, or as interactive 3D renderings~\cite{major2022tensorboard3d}. These can be used in conjunction with the engines mentioned in the previous section.

\subsubsection{Traceable transformations}

As mentioned in Section~\ref{transforms_array_dict}, MONAI transforms can be applied to both arrays and dictionaries of data. When dictionaries are used, as well as storing the input metadata corresponding to each image, a stack of transforms that have been applied to that image can also be stored. This is particularly useful when data augmentations are only applied some of the time, parameterized by user-defined probability (e.g., randomly flipping an image 50\% of the time), or where the user would like to inspect the parameters of a randomly applied transform (e.g., knowing the rotation matrix that was applied during \texttt{RandRotate}). It also provides a mechanism for augmentation auditing and traceability, a subject of increasing importance when developing AI models for clinical use. Internally, MONAI also uses this functionality when performing inverse transformations, when it is appropriate to invert with a last in, first out strategy.

\subsubsection{Interpretability}

\begin{figure}[!t]
  \centering
  \includegraphics[width=0.6\linewidth, angle=270]{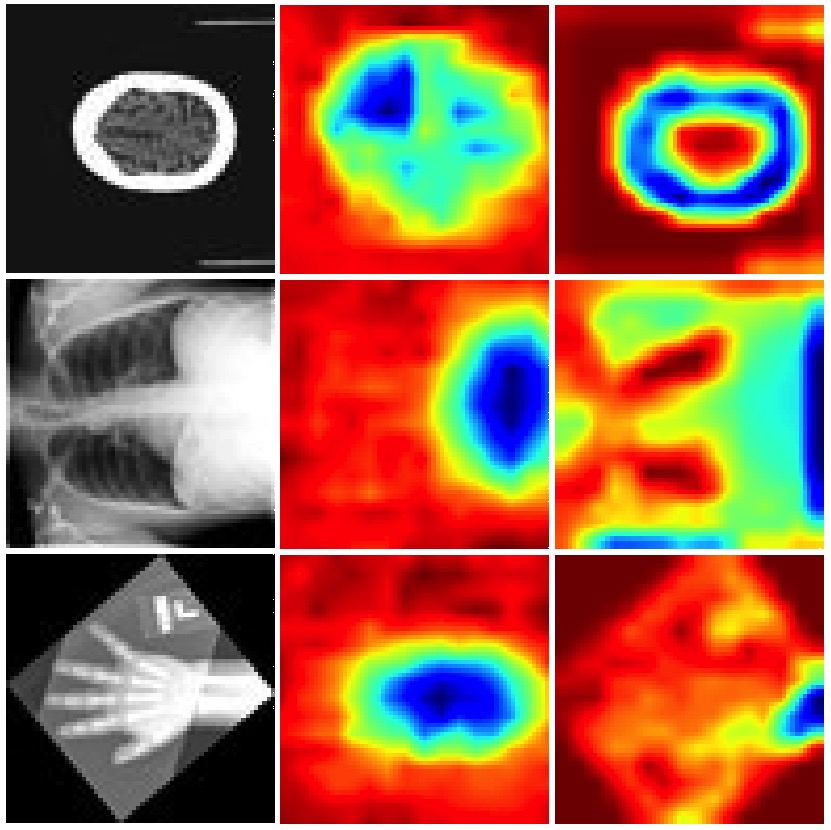}
  \caption{Examples of interpretability with MONAI. The MedNIST dataset was used here, and the network correctly classified each image, which from left to right were head CT, chest x-ray and hand x-ray. From top to bottom, the columns are the input image and then the outputs from occlusion sensitivity and GradCAM++ (a GradCAM variant). Blue indicates areas of the image that the network relied on more heavily in the decision-making process.}
  \label{fig:interpretability}
\end{figure}

MONAI supports popular variants of network interpretability that are often employed in classification tasks. These include occlusion sensitivity~\cite{OccSens}, GradCAM~\cite{GradCAM} and Smoothgrad~\cite{SmoothGrad}. These methods help the researchers address the question: ``which parts of the image were important in arriving to a given decision of a classification?''. This allows users to better visualize the inner workings of their models. Examples of occlusion sensitivity and GradCAM++ (a GradCAM variant) can be seen in Figure~\ref{fig:interpretability}.

Another form of interpretability is test-time augmentation (\gls*{TTA}), which is also available in MONAI. \gls*{TTA} works by repeatedly applying data augmentations and averaging the inferred outputs. If spatial transformations have been applied, the inverse transformation is required --- as previously mentioned, this functionality is also provided by MONAI.

\subsubsection{Utility functions}
Lastly, a wealth of other functions exist in MONAI for helping users visualize their results. An example of this is the \texttt{blend\_image} function, which allows users to easily create RGB images from the superposition of images and labels (ground truth or inferred), examples of which can be seen in Figure~\ref{fig:transforms}.
Another example of utility functions used for visualizations is the \texttt{matshow3d} function, which creates a 3D volume figure as a grid of images.

\section{Experiments and Applications}\label{Applications}
This section introduces four applications with the aim of demonstrating the breadth and benefits of MONAI. 

\subsection{Segmentation}

The typical PyTorch training workflow consists of a training loop which feeds data into the network and optimizes its parameters, followed by an evaluation loop where validation data is used in conjunction with a metric to assess training progress. 
PyTorch provides the relatively low-level components for this workflow but relies on libraries such as Ignite for higher-level definitions.
Types which encapsulate the training process offer ease of use at the expense of some flexibility but represent a significant acceleration of development for most users. 

Training a segmentation network with ground truth annotated data is one instance of this common supervised process.
The network is tasked with predicting the segmentation image from the input image, and this is compared using a loss such as Dice against the known actual segmentation. 
A supervised segmentation training workflow in PyTorch can be sketched as the following:

\begin{center}
\begin{tcolorbox}[
    enhanced,
    attach boxed title to top center={yshift=-2mm},
    colback=darkspringgreen!10,
    colframe=darkspringgreen,
    colbacktitle=darkspringgreen,
    title=PyTorch Training Loop,
    fonttitle=\bfseries\color{white},
    boxed title style={size=small,colframe=darkspringgreen,sharp corners},
    sharp corners,
]
\begin{minted}[fontsize=\small]{python}
for epoch in range(max_epochs):
    network.train()    
    for inputs, labels in train_loader:
        optimizer.zero_grad()
        outputs = network(inputs)
        loss = loss_function(outputs, labels)
        loss.backward()
        optimizer.step()

    network.eval()
    with torch.no_grad():
        for val_inputs, val_labels in val_loader:
            val_outputs = network(val_inputs)
            metric(y_pred=val_outputs, y=val_labels)

        metric = metric.aggregate().item()
        print("Validation result:", metric)
\end{minted}
\end{tcolorbox}
\end{center}

MONAI provides types inheriting from Ignite definitions to encapsulate this process. 
The purpose is to simplify and regularize how workflows are created to make the process quicker, easier, and more reproducible. 
For a supervised training workflow, the \texttt{SupervisedTrainer} and \texttt{SupervisedEvaluator} types define the training and validation components discussed previously.
The previous training script is thus transformed into an instantiation of these types followed by a call to \texttt{run} to hand control of the process to the internal training mechanism:

\begin{center}
\begin{tcolorbox}[
    enhanced,
    attach boxed title to top center={yshift=-2mm},
    colback=darkspringgreen!10,
    colframe=darkspringgreen,
    colbacktitle=darkspringgreen,
    title=MONAI Training Loop,
    fonttitle=\bfseries\color{white},
    boxed title style={size=small,colframe=darkspringgreen,sharp corners},
    sharp corners,
]
\begin{minted}[fontsize=\small]{python}
evaluator = SupervisedEvaluator(
    val_data_loader=val_loader,
    network=network,
    key_val_metric={ "metric": metric },
    ...
) 

trainer = SupervisedTrainer(
    max_epochs=num_epochs,
    train_data_loader=train_loader,
    network=network,
    optimizer=optimizer,
    loss_function=loss_function,
    train_handlers=[ValidationHandler(1,evaluator)],
    ...
)

trainer.run()  # do the training run for 10 epochs
\end{minted}
\end{tcolorbox}
\end{center}

In addition to simplifying the code down to a few lines of object instantiations, this form of workflow aids in reproducibility and correctness by limiting the number of details and definitions the scientist needs to implement for training.
With more moving parts and structural definitions a user must implement, even simple things like loops and method calls, the more opportunity there is for variation or error to be introduced into the code.
Reproduction for software experiments requires rerunning or repurposing pre-existing code for new situations, and so introduced variations between code bases that are expected to be equivalent complicate the process of deriving and verifying scientific results.

\subsection{Classification}

For many training and evaluation tasks, deterministic behavior is important for consistent and reproducible results. 
MONAI provides routines for managing determinism in workflows by setting random seed values for PyTorch, Numpy, and the native Python random library, as well as setting backend flags, to appropriate values.
This ensures that many stochastic processes, including the evaluation of data transforms and network processes like dropout, can be expected to produce identical results between runs.

For example, a fixed seed can be set for all involved libraries as well as setting flags for deterministic behavior, followed by computation, and then returning the state of the running program to non-determinism:

\begin{center}
\begin{tcolorbox}[
    enhanced,
    attach boxed title to top center={yshift=-2mm},
    colback=darkspringgreen!10,
    colframe=darkspringgreen,
    colbacktitle=darkspringgreen,
    title=MONAI Training Loop,
    fonttitle=\bfseries\color{white},
    boxed title style={size=small,colframe=darkspringgreen,sharp corners},
    sharp corners,
]
\begin{minted}[fontsize=\small]{python}
random_seed = 42
monai.utils.set_determinism(
    seed=random_seed, 
    use_deterministic_algorithms=True
)

# perform computation

monai.utils.set_determinism(
    seed=None, 
    use_deterministic_algorithms=False
)
\end{minted}
\end{tcolorbox}
\end{center}

Classification problems benefit from this determinism when interpreting network behavior and results. 
MONAI provides networks, losses, and metrics for classification problems, and visualization routines for displaying occlusion sensitivity and the GradCAM method.
Making sense of the interpretations these methods provide requires experimentation and minor changes to network parameters to observe results.
Being able to fix a deterministic training process allows the user to minimize the variability that would be introduced with randomized transforms or internal PyTorch/Numpy processes in their code, and allows consistent results to be generated and visualized.

\subsection{Registration}

DeepReg~\cite{Fu2020} is an open-source toolkit of networks and algorithms for image registration.
A number of these components have been ported into MONAI and are provided as standard loss functions and network definitions. 

MONAI provides a semi-supervised example of registering lung CT images at inspiration to expiration.
During training, a deformation field is estimated by the network to deform the moving image and its segmentation to the fixed image and segmentation.
A combined loss scores the results based on the similarity between the deformed and fixed images, the Dice score of the deformed and fixed segmentation calculated at various scales, and an energy-based regularisation term.

The image loss is the standard PyTorch MSE function, and the Dice loss is that provided by MONAI and used in other segmentation applications.
The regularisation function is a bending loss term based on second-order derivatives and is a MONAI definition.

Deformation is performed using a MONAI class which can use resampling routines included in PyTorch, or MONAI's compiled code for doing the same.
Note that these routines implement interpolation models (e.g. tricubic), which are not present in PyTorch, in addition to having speed advantages. 
MONAI thus provides a number of important tools for training deep learning registration algorithms not present in the base PyTorch library.

\subsection{Benchmarks}

Typically, model training is a time-consuming step during deep learning development, especially in medical imaging applications. Volumetric medical images are usually large (as multi-dimensional arrays), and the model training process can be complex. Even with powerful hardware platforms (e.g., CPU/GPU with large RAM), it is not easy to fully leverage them to achieve high performance. MONAI provides not only optimized tools and code but also a detailed guide based on best practices to achieve the best performance. 

NVIDIA GPUs have been widely applied in many areas of deep learning training and evaluation, and the CUDA parallel computation library shows obvious acceleration when compared to traditional computation methods. To fully leverage GPU features, many popular mechanisms are provided in PyTorch, like automatic mixed precision (AMP), distributed data-parallel, etc. MONAI can support these features and provides rich examples. Even with \texttt{CacheDataset}, one still needs to copy the same data to GPU memory for random transforms or network computation for every epoch. To address this issue, an efficient approach is to cache the data to GPU memory directly, allowing for more efficient and immediate GPU computation. 

As an example, one first converts to PyTorch Tensor with the \texttt{EnsureTyped} transform and moves data to GPU with the \texttt{ToDeviced} transform. \texttt{CacheDataset} then caches the transform results until \texttt{ToDeviced}, so caches are stored in GPU memory. Then in every epoch, the program fetches cached data from GPU memory and only executes the following random transforms (like \texttt{RandCropByPosNegLabeld}) on GPU directly. 

A fast model training guide is available as a MONAI tutorial, providing an overview of the fast training techniques in practice, covering aspects such as profiling the pipelines, optimizing data loading function, algorithmic improvement, optimizing GPU utilization, and leveraging multi-GPU and multi-node distributed training. 

\section{Conclusions}
We present MONAI, an open-source and community-supported, consortium-led PyTorch-based framework for deep learning in healthcare, with a particular focusing providing a standardized and flexible platform for developing biomedical and clinical applications. 
This paper introduces key design considerations and implementation choices that have driven the development of MONAI, presents the rationale behind the MONAI open-source strategy and license, and introduces MONAI's major system components. Finally, example use cases of MONAI are provided, highlighting the benefits of the platform. 
Overall, we believe MONAI can help accelerate and simplify the development of AI models and contribute towards more impactful research, development, and clinical deployment.

\section{Acknowledgments}
JC, EK, TV, and SO acknowledge funding from the Wellcome/EPSRC Centre for Medical Engineering (WT203148/Z/16/Z), Wellcome Flagship Programme (WT213038/Z/18/Z). JC, EK, SO and HS also acknowledge funding from the London Medical Imaging and AI Centre for Value-based Healthcare.
SA was funded, in part, from the National Institutes of Health via NIBIB and NIGMS R01EB021396, NIBIB R01EB014955, NCI R01CA220681, and NINDS R42NS086295. This project has been funded in whole or in part with federal funds from the National Cancer Institute, National Institutes of Health, under Contract No. HHSN261200800001E. The content of this publication does not necessarily reflect the views or policies of the Department of Health and Human Services, nor does mention of trade names, commercial products, or organizations imply endorsement by the U.S. Government.

\backmatter

\noindent

\bibliography{sn-bibliography} 

\begin{thebibliography}{10}
\expandafter\ifx\csname url\endcsname\relax
  \def\url#1{\burl{#1}}\fi
\expandafter\ifx\csname urlprefix\endcsname\relax\def\urlprefix{URL }\fi
\providecommand{\bibinfo}[2]{#2}
\providecommand{\eprint}[2][]{\url{#2}}
\providecommand{\doi}[1]{\url{https://doi.org/#1}}
\bibcommenthead

\bibitem{Kraljevic:2021aa}
\bibinfo{author}{Kraljevic, Z.} \emph{et~al.}
\newblock \bibinfo{title}{Multi-domain clinical natural language processing
  with medcat: The medical concept annotation toolkit}.
\newblock \emph{\bibinfo{journal}{Artificial Intelligence in Medicine}}
  \textbf{\bibinfo{volume}{117}}, \bibinfo{pages}{102083}
  (\bibinfo{year}{2021}).
\newblock
  \urlprefix\url{https://www.sciencedirect.com/science/article/pii/S0933365721000762}.
\newblock \doi{https://doi.org/10.1016/j.artmed.2021.102083} .

\bibitem{Brzezicki:2020aa}
\bibinfo{author}{Brzezicki, M.~A.} \emph{et~al.}
\newblock \bibinfo{title}{Artificial intelligence outperforms human students in
  conducting neurosurgical audits}.
\newblock \emph{\bibinfo{journal}{Clinical Neurology and Neurosurgery}}
  \textbf{\bibinfo{volume}{192}}, \bibinfo{pages}{105732}
  (\bibinfo{year}{2020}).
\newblock
  \urlprefix\url{https://www.sciencedirect.com/science/article/pii/S0303846720300755}.
\newblock \doi{https://doi.org/10.1016/j.clineuro.2020.105732} .

\bibitem{Nelson:2019aa}
\bibinfo{author}{Nelson, A.}, \bibinfo{author}{Herron, D.},
  \bibinfo{author}{Rees, G.} \& \bibinfo{author}{Nachev, P.}
\newblock \bibinfo{title}{Predicting scheduled hospital attendance with
  artificial intelligence}.
\newblock \emph{\bibinfo{journal}{npj Digital Medicine}}
  \textbf{\bibinfo{volume}{2}}~(1), \bibinfo{pages}{26} (\bibinfo{year}{2019}).
\newblock \urlprefix\url{https://doi.org/10.1038/s41746-019-0103-3}.
\newblock \doi{10.1038/s41746-019-0103-3} .

\bibitem{Panch:2019aa}
\bibinfo{author}{Panch, T.}, \bibinfo{author}{Mattie, H.} \&
  \bibinfo{author}{Celi, L.~A.}
\newblock \bibinfo{title}{The ``inconvenient truth''about ai in healthcare}.
\newblock \emph{\bibinfo{journal}{npj Digital Medicine}}
  \textbf{\bibinfo{volume}{2}}~(1), \bibinfo{pages}{77} (\bibinfo{year}{2019}).
\newblock \urlprefix\url{https://doi.org/10.1038/s41746-019-0155-4}.
\newblock \doi{10.1038/s41746-019-0155-4} .

\bibitem{tensorflowpaper}
\bibinfo{author}{Abadi, M.} \emph{et~al.}
\newblock \bibinfo{title}{Tensorflow: A system for large-scale machine
  learning} (\bibinfo{year}{2016}).
\newblock
  \urlprefix\url{https://www.usenix.org/system/files/conference/osdi16/osdi16-abadi.pdf}.

\bibitem{pytorchpaper}
\bibinfo{author}{Paszke, A.} \emph{et~al.}
\newblock \bibinfo{title}{Pytorch: An imperative style, high-performance deep
  learning library} (\bibinfo{year}{2019}).
\newblock \urlprefix\url{https://arxiv.org/abs/1912.01703}.

\bibitem{mxnetpaper}
\bibinfo{author}{Chen, T.} \emph{et~al.}
\newblock \bibinfo{title}{Mxnet: A flexible and efficient machine learning
  library for heterogeneous distributed systems} (\bibinfo{year}{2015}).
\newblock \urlprefix\url{https://arxiv.org/abs/1512.01274}.

\bibitem{Gibson:2018aa}
\bibinfo{author}{Gibson, E.} \emph{et~al.}
\newblock \bibinfo{title}{Niftynet: a deep-learning platform for medical
  imaging}.
\newblock \emph{\bibinfo{journal}{Computer Methods and Programs in
  Biomedicine}} \textbf{\bibinfo{volume}{158}}, \bibinfo{pages}{113--122}
  (\bibinfo{year}{2018}).
\newblock
  \urlprefix\url{https://www.sciencedirect.com/science/article/pii/S0169260717311823}.
\newblock \doi{https://doi.org/10.1016/j.cmpb.2018.01.025} .

\bibitem{DLTKpaper}
\bibinfo{author}{Pawlowski, N.} \emph{et~al.}
\newblock \bibinfo{title}{Dltk: State of the art reference implementations for
  deep learning on medical images} (\bibinfo{year}{2017}).
\newblock \urlprefix\url{https://arxiv.org/abs/1711.06853}.

\bibitem{deepneuropaper}
\bibinfo{author}{Beers, A.} \emph{et~al.}
\newblock \bibinfo{title}{Deepneuro: an open-source deep learning toolbox for
  neuroimaging.}
\newblock \emph{\bibinfo{journal}{Neuroinformatics}}
  \textbf{\bibinfo{volume}{19}}~(1), \bibinfo{pages}{127--140}
  (\bibinfo{year}{2021}).
\newblock \doi{10.1007/s12021-020-09477-5} .

\bibitem{he2019mlframeworks}
\bibinfo{author}{He, H.}
\newblock \bibinfo{title}{The state of machine learning frameworks in 2019}.
\newblock \emph{\bibinfo{journal}{The Gradient}}  (\bibinfo{year}{2019}) .

\bibitem{MONAILabel}
\bibinfo{author}{Diaz-Pinto, A.} \emph{et~al.}
\newblock \bibinfo{title}{Monai label: A framework for ai-assisted interactive
  labeling of 3d medical images} (\bibinfo{year}{2022}).
\newblock \urlprefix\url{https://arxiv.org/abs/2203.12362}.

\bibitem{itk2014}
\bibinfo{author}{McCormick, M.}, \bibinfo{author}{Liu, X.},
  \bibinfo{author}{Ibanez, L.}, \bibinfo{author}{Jomier, J.} \&
  \bibinfo{author}{Marion, C.}
\newblock \bibinfo{title}{{ITK: enabling reproducible research and open
  science}}.
\newblock \emph{\bibinfo{journal}{Frontiers in Neuroinformatics}}
  \textbf{\bibinfo{volume}{8}}, \bibinfo{pages}{13} (\bibinfo{year}{2014}).
\newblock
  \urlprefix\url{https://www.frontiersin.org/article/10.3389/fninf.2014.00013}.
\newblock \doi{10.3389/fninf.2014.00013} .

\bibitem{nibabel}
\bibinfo{author}{Brett, M.} \emph{et~al.}
\newblock \bibinfo{title}{nipy/nibabel: 3.2.1} (\bibinfo{year}{2020}).
\newblock \urlprefix\url{https://doi.org/10.5281/zenodo.4295521}.

\bibitem{PIL}
\bibinfo{author}{van Kemenade, H.} \emph{et~al.}
\newblock \bibinfo{title}{python-pillow/pillow: 8.4.0} (\bibinfo{year}{2021}).
\newblock \urlprefix\url{https://doi.org/10.5281/zenodo.5571504}.

\bibitem{testtimeaugmentation2019}
\bibinfo{author}{Wang, G.} \emph{et~al.}
\newblock \bibinfo{title}{{Aleatoric uncertainty estimation with test-time
  augmentation for medical image segmentation with convolutional neural
  networks}}.
\newblock \emph{\bibinfo{journal}{Neurocomputing}}
  \textbf{\bibinfo{volume}{338}}, \bibinfo{pages}{34--45}
  (\bibinfo{year}{2019}).
\newblock
  \urlprefix\url{https://www.sciencedirect.com/science/article/pii/S0925231219301961}.
\newblock \doi{https://doi.org/10.1016/j.neucom.2019.01.103} .

\bibitem{perez-garcia_torchio_2021}
\bibinfo{author}{P{\'e}rez-Garc{\'\i}a, F.}, \bibinfo{author}{Sparks, R.} \&
  \bibinfo{author}{Ourselin, S.}
\newblock \bibinfo{title}{Torchio: a python library for efficient loading,
  preprocessing, augmentation and patch-based sampling of medical images in
  deep learning}.
\newblock \emph{\bibinfo{journal}{Computer Methods and Programs in
  Biomedicine}} \bibinfo{pages}{106236} (\bibinfo{year}{2021}).
\newblock
  \urlprefix\url{https://www.sciencedirect.com/science/article/pii/S0169260721003102}.
\newblock \doi{https://doi.org/10.1016/j.cmpb.2021.106236} .

\bibitem{eriba2019kornia}
\bibinfo{author}{Riba, E.}, \bibinfo{author}{Mishkin, D.},
  \bibinfo{author}{Ponsa, D.}, \bibinfo{author}{Rublee, E.} \&
  \bibinfo{author}{Bradski, G.}
\newblock \bibinfo{title}{Kornia: an open source differentiable computer vision
  library for pytorch} (\bibinfo{year}{2020}).
\newblock \urlprefix\url{https://arxiv.org/pdf/1910.02190.pdf}.

\bibitem{isensee_fabian_2020_3632567}
\bibinfo{author}{Isensee, F.} \emph{et~al.}
\newblock \bibinfo{title}{{batchgenerators - a python framework for data
  augmentation}} (\bibinfo{year}{2020}).
\newblock \urlprefix\url{https://doi.org/10.5281/zenodo.3632567}.

\bibitem{rising}
\bibinfo{author}{Schock, J.}, \bibinfo{author}{Baumgartner, M.} \&
  \bibinfo{author}{Weninger, L.}
\newblock \bibinfo{title}{Phoenixdl/rising: High-performance differentiable
  medical data augmentation}.
\newblock \bibinfo{howpublished}{\url{https://github.com/PhoenixDL/rising}}.
\newblock \bibinfo{note}{Accessed: 2021-12-20}.

\bibitem{cuCIM}
\bibinfo{author}{Lee, G.}, \bibinfo{author}{Bae, G.}, \bibinfo{author}{Zaitlen,
  B.}, \bibinfo{author}{Kirkham, J.} \& \bibinfo{author}{Choudhury, R.}
\newblock \bibinfo{title}{cucim - a gpu image i/o and processing library}
  (\bibinfo{year}{2021}).
\newblock \urlprefix\url{https://doi.org/10.25080/majora-1b6fd038-022}.

\bibitem{Sudre_2017}
\bibinfo{author}{Sudre, C.~H.}, \bibinfo{author}{Li, W.},
  \bibinfo{author}{Vercauteren, T.}, \bibinfo{author}{Ourselin, S.} \&
  \bibinfo{author}{Jorge~Cardoso, M.}
\newblock \bibinfo{editor}{Cardoso, M.~J.} \emph{et~al.} (eds)
  \emph{\bibinfo{title}{Generalised dice overlap as a deep learning loss
  function for highly unbalanced segmentations}}.
\newblock (eds \bibinfo{editor}{Cardoso, M.~J.} \emph{et~al.})
  \emph{\bibinfo{booktitle}{Deep Learning in Medical Image Analysis and
  Multimodal Learning for Clinical Decision Support}},
  \bibinfo{pages}{240--248} (\bibinfo{publisher}{Springer International
  Publishing}, \bibinfo{address}{Cham}, \bibinfo{year}{2017}).

\bibitem{MaierHein2022}
\bibinfo{author}{Maier-Hein, L.} \emph{et~al.}
\newblock \bibinfo{title}{Metrics reloaded: Pitfalls and recommendations for
  image analysis validation} (\bibinfo{year}{2022}).
\newblock \urlprefix\url{https://arxiv.org/abs/2206.01653}.

\bibitem{Reinke2021}
\bibinfo{author}{Reinke, A.} \emph{et~al.}
\newblock \bibinfo{title}{Common limitations of image processing metrics: A
  picture story} (\bibinfo{year}{2021}).
\newblock \urlprefix\url{https://arxiv.org/abs/2104.05642}.

\bibitem{Ronneberger2015}
\bibinfo{author}{Ronneberger, O.}, \bibinfo{author}{Fischer, P.} \&
  \bibinfo{author}{Brox, T.}
\newblock \bibinfo{title}{ in \textit{{U-Net: Convolutional Networks for
  Biomedical Image Segmentation}}} (eds \bibinfo{editor}{Navab, N.},
  \bibinfo{editor}{Hornegger, J.}, \bibinfo{editor}{{Wells III}, W.~M.} \&
  \bibinfo{editor}{Frangi, A.~F.}) \emph{\bibinfo{booktitle}{MICCAI 2015}},
  Vol. \bibinfo{volume}{9351} of \emph{\bibinfo{series}{LNCS}}
  \bibinfo{pages}{234--241} (\bibinfo{publisher}{Springer},
  \bibinfo{year}{2015}).

\bibitem{https://doi.org/10.48550/arxiv.1502.01852}
\bibinfo{author}{He, K.}, \bibinfo{author}{Zhang, X.}, \bibinfo{author}{Ren,
  S.} \& \bibinfo{author}{Sun, J.}
\newblock \bibinfo{title}{Delving deep into rectifiers: Surpassing human-level
  performance on imagenet classification} (\bibinfo{year}{2015}).
\newblock \urlprefix\url{https://arxiv.org/abs/1502.01852}.

\bibitem{gelu}
\bibinfo{author}{Hendrycks, D.} \& \bibinfo{author}{Gimpel, K.}
\newblock \bibinfo{title}{Gaussian error linear units (gelus)}
  (\bibinfo{year}{2016}).
\newblock \urlprefix\url{https://arxiv.org/abs/1606.08415}.

\bibitem{https://doi.org/10.48550/arxiv.1706.02515}
\bibinfo{author}{Klambauer, G.}, \bibinfo{author}{Unterthiner, T.},
  \bibinfo{author}{Mayr, A.} \& \bibinfo{author}{Hochreiter, S.}
\newblock \bibinfo{title}{Self-normalizing neural networks}
  (\bibinfo{year}{2017}).
\newblock \urlprefix\url{https://arxiv.org/abs/1706.02515}.

\bibitem{lecun2010mnist}
\bibinfo{author}{LeCun, Y.}, \bibinfo{author}{Cortes, C.} \&
  \bibinfo{author}{Burges, C.}
\newblock \bibinfo{title}{Mnist handwritten digit database}.
\newblock \emph{\bibinfo{journal}{ATT Labs [Online]. Available:
  http://yann.lecun.com/exdb/mnist}} \textbf{\bibinfo{volume}{2}}
  (\bibinfo{year}{2010}) .

\bibitem{decathlon}
\bibinfo{author}{Antonelli, M.} \emph{et~al.}
\newblock \bibinfo{title}{The medical segmentation decathlon}
  (\bibinfo{year}{2021}).
\newblock \urlprefix\url{https://arxiv.org/abs/2106.05735}.

\bibitem{abadi2016tensorflow}
\bibinfo{author}{Abadi, M.} \emph{et~al.}
\newblock \bibinfo{title}{Tensorflow: Large-scale machine learning on
  heterogeneous distributed systems} (\bibinfo{year}{2016}).
\newblock \eprint{1603.04467}.

\bibitem{major2022tensorboard3d}
\bibinfo{author}{Major, B.}, \bibinfo{author}{McCormick, M.} \&
  \bibinfo{author}{Aylward, S.}
\newblock \bibinfo{title}{Tensorboardplugin3d: 3d tensor visualization}.
\newblock
  \bibinfo{howpublished}{\url{https://github.com/KitwareMedical/tensorboard-plugin-3d}}
  (\bibinfo{year}{2022}).

\bibitem{OccSens}
\bibinfo{author}{Zeiler, M.~D.} \& \bibinfo{author}{Fergus, R.}
\newblock \bibinfo{editor}{Fleet, D.}, \bibinfo{editor}{Pajdla, T.},
  \bibinfo{editor}{Schiele, B.} \& \bibinfo{editor}{Tuytelaars, T.} (eds)
  \emph{\bibinfo{title}{{Visualizing and Understanding Convolutional
  Networks}}}.
\newblock (eds \bibinfo{editor}{Fleet, D.}, \bibinfo{editor}{Pajdla, T.},
  \bibinfo{editor}{Schiele, B.} \& \bibinfo{editor}{Tuytelaars, T.})
  \emph{\bibinfo{booktitle}{Computer Vision -- ECCV 2014}},
  \bibinfo{pages}{818--833} (\bibinfo{publisher}{Springer International
  Publishing}, \bibinfo{address}{Cham}, \bibinfo{year}{2014}).

\bibitem{GradCAM}
\bibinfo{author}{Selvaraju, R.~R.} \emph{et~al.}
\newblock \bibinfo{title}{Grad-cam: Visual explanations from deep networks via
  gradient-based localization} (\bibinfo{year}{2017}).

\bibitem{SmoothGrad}
\bibinfo{author}{Smilkov, D.}, \bibinfo{author}{Thorat, N.},
  \bibinfo{author}{Kim, B.}, \bibinfo{author}{Vi{\'e}gas, F.} \&
  \bibinfo{author}{Wattenberg, M.}
\newblock \bibinfo{title}{Smoothgrad: removing noise by adding noise}
  (\bibinfo{year}{2017}).
\newblock \urlprefix\url{https://arxiv.org/abs/1706.03825}.

\bibitem{Fu2020}
\bibinfo{author}{Fu, Y.} \emph{et~al.}
\newblock \bibinfo{title}{{DeepReg}: a deep learning toolkit for medical image
  registration}.
\newblock \emph{\bibinfo{journal}{Journal of Open Source Software}}
  \textbf{\bibinfo{volume}{5}}~(55), \bibinfo{pages}{2705}
  (\bibinfo{year}{2020}).
\newblock \urlprefix\url{https://doi.org/10.21105/joss.02705}.
\newblock \doi{10.21105/joss.02705} .

\end{thebibliography}

\end{document}